\documentclass{article}

\usepackage[final]{timeseries_workshop}
\usepackage{enumitem}
\usepackage{tikz}
\newcommand*\circled[1]{\tikz[baseline=(char.base)]{
            \node[shape=circle,draw,inner sep=0.5pt] (char) {#1};}}

\usepackage[utf8]{inputenc} 
\usepackage[T1]{fontenc}    
\usepackage{hyperref}       
\usepackage{url}            
\usepackage{booktabs}       
\usepackage{amsfonts}       
\usepackage{nicefrac}       
\usepackage{microtype}      
\usepackage{xcolor}         
\usepackage{graphicx}
\usepackage[frozencache,cachedir=./minted]{minted}

\usepackage{mdframed}
\usepackage{multicol}
\usepackage{subcaption}
\usepackage{xspace}
\usepackage{multirow}
\usepackage{enumitem}

\def\systemName{\texttt{\textbf{\large FMTK}}\xspace}
\definecolor{myblue}{HTML}{254E70}
\usepackage{titlesec}
\titlespacing*{\section}
  {0pt}
  {1.5ex plus 0.1ex minus .2ex}
  {0.1ex}                

\titlespacing*{\subsection}
  {0pt}
  {1.5ex plus .01ex minus .2ex} 
  {0.1ex}

\begin{document}
\title{FMTK: A Modular Toolkit for Composable Time Series Foundation Model Pipelines}
\author{
    Hetvi Shastri\textsuperscript{1} \quad
    Pragya Sharma\textsuperscript{2} \quad
    Walid A. Hanafy\textsuperscript{1} \quad
    Mani Srivastava\textsuperscript{2\dag} \quad
    Prashant Shenoy\textsuperscript{1} \\
    \textsuperscript{1}University of Massachusetts Amherst \\
    \textsuperscript{2}University of California Los Angeles
}

\maketitle
\begingroup
\renewcommand\thefootnote{\dag}
\footnotetext{Author holds concurrent appointments as a Professor at UCLA, and as an Amazon Scholar. This paper describes work performed at UCLA and is not associated with Amazon.}
\endgroup
\vspace{-1em}
\begin{abstract}
Foundation models (FMs) have opened new avenues for machine learning applications due to their ability to adapt to new and unseen tasks with minimal or no further training. Time-series foundation models (TSFMs)---FMs trained on time-series data---have shown strong performance on classification, regression, and imputation tasks. Recent pipelines combine TSFMs with task-specific encoders, decoders, and adapters to improve performance; however, assembling such pipelines typically requires ad hoc, model-specific implementations that hinder modularity and reproducibility. We introduce \systemName, an open-source, lightweight and extensible toolkit for constructing and fine-tuning TSFM pipelines via standardized backbone and component abstractions. \systemName enables flexible composition across models and tasks, achieving correctness and performance with an average of seven lines of code.~\url{https://github.com/umassos/FMTK}
    
\end{abstract}

\vspace{-1em}
\section{Introduction}
Time Series Foundation Models (TSFMs), such as MOMENT~\cite{Goswami2024:MOMENT}, Chronos~\cite{Ansari2024:Chronos}, and TimesFM~\cite{Das2024:TimesFM}, have emerged as powerful pre-trained architectures for a variety of downstream tasks, including forecasting, classification, and imputation. While these models serve as fixed backbones trained on large-scale temporal data, effective task specialization often requires integrating additional components: input encoders to structure raw data, task-specific decoders to generate predictions, and increasingly, parameter-efficient adapters (e.g., LoRA~\cite{Hu2022:LoRA}) to enable lightweight fine-tuning.

This modular design space, though conceptually flexible, has resulted in a fragmented and ad hoc implementation landscape. For example, models such as PaPaGei~\cite{pillai2024papagei} and Mantis~\cite{feofanov2025mantis} require the development of extensive custom pipelines for simple comparison with state-of-the-art models. Moreover, models like Moment~\cite{Goswami2024:MOMENT} offer distinct modes for different tasks (e.g., forecasting versus classification), which restricts the reuse of the same powerful backbone for a diverse set of downstream tasks during runtime. As a result, three key challenges emerge. \circled{1} First, the absence of a unifying abstraction across encoders, backbones, adapters, and decoders significantly increases the engineering burden and inhibits systematic exploration of architectural variants. \circled{2} Second, the lack of modular encapsulation complicates the attribution. It becomes difficult to isolate and measure the contribution of individual components to the overall performance of the model. \circled{3} Third, evaluation practices vary widely across studies. Minor differences in data pre-processing, training regimes, or decoder heads can lead to substantial discrepancies in reported results, thereby undermining reproducibility.

Although existing time series libraries such as sktime~\cite{sktime}, Darts~\cite{Julien2022:Darts}, tsai~\cite{tsai}, and GluonTS~\cite{gluonts_jmlr} support classical and deep learning pipelines, they do not address the emerging need for composable, FM-centric evaluation. To this end, we introduce \systemName: an open-source, lightweight, and extensible Time Series Foundation Model Toolkit for constructing, fine-tuning and benchmarking modular TSFM pipelines.

\noindent\textbf{Contributions:} Our contributions can be summarized as follows:
\begin{enumerate}[leftmargin=*]
    \item \systemName proposes a standardized API for TSFM pipelines that defines a common grammar for connecting FM backbones with external encoders, fine-tuning adapters and decoders. 
    \item \systemName provides reference implementation of commonly used configurations and supports multiple time series tasks under consistent evaluation settings.
    \item By decoupling architectural components and enforcing standardized execution semantics, \systemName facilitates reproducible experimentation and controlled comparison across a rapidly growing space of TSFM-based systems. 
\end{enumerate}

\section{Design}
\begin{figure}[t!]
    \centering
    \includegraphics[width=\linewidth]{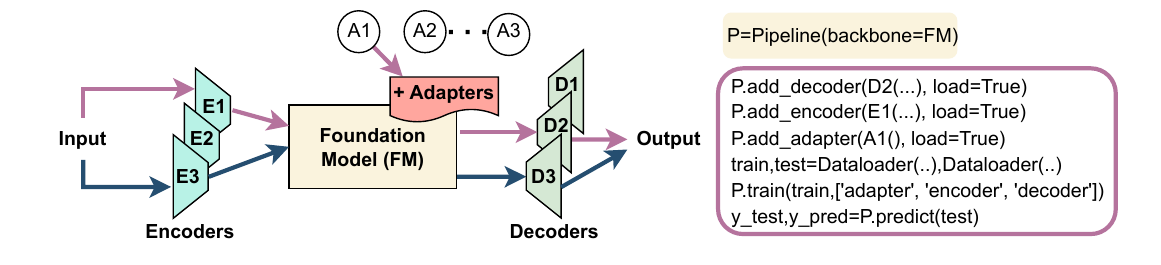}
    \caption{\textbf{Modular abstraction of pipeline construction using \systemName}: The framework allows instantiating pipelines by pairing an FM with interchangeable components. Users can dynamically select and load components, specify trainable parts (e.g., decoder), and benchmark pipelines in a unified interface. We illustrate two example configurations using the same FM: \textbf{\textcolor{purple}{(top)}} \color{black} encoder-decoder-adapter-tuned pipeline with E1, A1 and D2; \textbf{\textcolor{myblue}{(bottom)}} \color{black} encoder-decoder tuned with E3 and D3.}
    \label{fig:architecture}
    \vspace{-1em}
\end{figure}

To address the challenges outlined earlier (\circled{1}–\circled{3}), the design of \systemName is guided by three core principles: composability, usability, and reproducibility. \textit{Composability} is achieved through standardized interfaces for encoders, backbones, adapters, and decoders, enabling seamless interchangeability and systematic exploration of pipeline configurations. \textit{Usability} is prioritized through a lightweight and declarative API that abstracts away engineering complexity, making it possible for both domain experts and non-specialists to experiment with modular TSFM pipelines without extensive prior familiarity with FM internals. Finally, to promote \textit{reproducibility}, the toolkit enforces consistent pre-processing, component integration, evaluation routines, and monitoring capabilities, allowing researchers to conduct controlled comparisons and ablation studies under uniform conditions. 

\subsection{Components}

These design principles manifest in \systemName through a modular architecture centered around four key components as shown in  Figure~\ref{fig:architecture}. \textbf{Encoders} transform raw time series into representations compatible with the foundation model. These may perform format conversion, context windowing, dimentionality reduction or domain-specific preprocessing. \systemName allows interchangeable encoders to support diverse inputs.
\textbf{Foundation Model Backbone (FM)} refers to a pre-trained, frozen or partially frozen TSFM that produces task-agnostic latent features. \systemName supports FMs like Chronos, Moment, and TimesFM, by providing a common I/O interfaces.
\textbf{Adapters} are lightweight modules for efficient adaptation to new tasks or domains without full backbone fine-tuning. \systemName currently supports methods such as LoRA enabling selective training.
\textbf{Decoders} are task-specific heads (e.g., for forecasting, classification) that operate on FM outputs. They may range from simple MLPs to more complex attention-based or partially fine-tuned heads.

\subsection{Pipeline Abstraction}
To support flexible experimentation, \systemName exposes a unified \texttt{Pipeline} abstraction, which allows any valid combination of encoder, backbone, adapter, and decoder modules. As illustrated in Figure~\ref{fig:architecture}, different configurations traverse distinct paths through the component graph depending on the intended usage pattern. We demonstrate via two use-cases how the system handles component integration, fine-tuning and execution behind the scenes, allowing both novice users and expert researchers to build, train, and evaluate pipelines with minimal overhead. (More details provided in Appendix \ref{appA})

\textbf{Use-Case 1 (simple usage):} Chronos backbone can be combined with SVM decoder to enable Heartbeat Classification task. \systemName allows this by attaching the decoder through \texttt{add\_decoder()} and restricting training to the decoder via \texttt{train(parts\_to\_train=[...])}.

\textbf{Use-Case 2 (advanced usage):} 
The Moment backbone can support diverse downstream tasks, such as PPG monitoring and Energy Forecasting, through a shared representation. PPG monitoring is supported by attaching a regression MLP decoder, linear encoder, and LoRA adapter to the backbone, while Energy Forecasting is supported by attaching an MLP decoder alone. After fine-tuning as explained in Use-Case 1, toolkit enables adaptive switching between distinct paths at runtime via
\texttt{load\_encoder()}, \texttt{load\_adapter()} and \texttt{load\_decoder()}.

\subsection{Runtime Metrics and Benchmarking Support}
In addition to traditional accuracy-based or MAE-based evaluation, \systemName supports the collection of runtime metrics for pipeline benchmarking. For each experiment, the framework logs:
\begin{enumerate}[leftmargin=*, label=\arabic*.]
    \item \textbf{Memory Utilization}: Peak GPU/CPU memory usage during inference and training.
    \item \textbf{Training and Inference Time}: Wall-clock time for each training phase and prediction batch.
    \item \textbf{Component Loading Overhead}: Time to load and switch between components at runtime.
    \item \textbf{Energy Consumption}: Optional support for energy profiling using compatible hardware monitors.
\end{enumerate}

These measurements facilitate holistic comparisons between pipeline variants and promote reproducibility by modularizing system-level bottlenecks and trade-offs.

\begin{table}[!tp]
\centering
\scriptsize
\begin{tabular}{llcccccc}
\toprule
                                 &                        & \multicolumn{2}{c}{\textbf{Chronos}} & \multicolumn{2}{c}{\textbf{Moment}} & \multicolumn{2}{c}{\textbf{PaPaGei-S}} \\ \cmidrule{3-8} 
\textbf{Task}                    & \textbf{Decoder}       & Base              & FMTK             & Base             & FMTK             & Base               & FMTK              \\ \toprule
\textbf{Regression (MAE)}        &                        &                   &                  &                  &                  &                    &                   \\
Systolic BP (PPG-BP)             & \multirow{3}{*}{Ridge} & 15.84             & 15.82            & 15.99            & 15.99            & 15.65              & 15.65             \\
Diastolic BP (PPG-BP)            &                        & 9.43              & 9.56             & 8.88             & 8.88             & 8.98               & 8.98              \\
Heart Rate (PPG-BP)              &                        & 9.04              & 9.04             & 5.24             & 5.24             & 6.32               & 6.32              \\ \midrule

\textbf{Classification (Acc \%)} &                        &                   &                  &                  &                  &                    &                   \\
Heartbeat (ECG5000)              & SVM                    & --                & 93.15            & 93.42            & 94.02            & --                 & 89.88            \\ \midrule
\textbf{Forecasting (MAE)}       &                        &                   &                  &                  &                  &                    &                   \\
Energy (ETTh1)                   & MLP                    & --                & 0.76             & 0.43             & 0.54             & --                 & 0.83             \\
\bottomrule
\end{tabular}
\vspace{1em}\caption{MAE and Accuracy comparison between baseline and \systemName across regression, classification, and forecasting tasks using Chronos-large, Moment-large, and PaPaGei-S backbones with Ridge, SVM and MLP decoder.}
\label{table:accuracy_comparison}
\vspace{-2em}
\end{table}

\section{Evaluation}
We implement \systemName in approximately 1.6k lines of Python code, leveraging PyTorch 2.7.1 for model execution and integration with widely used transformer and fine-tuning libraries~\cite{peft}. All experiments are conducted on a Linux system equipped with an NVIDIA A100 GPU, using Python 3.10.18 and CUDA 12.6. FM backbones are sourced from publicly available repositories: Chronos and Moment are retrieved via HuggingFace, while Papagei is initialized from locally downloaded checkpoints. We have performed three types of tasks regression, classification and forecating using PPG-BP~\cite{Liang2018ppgbp}, ECG5000~\cite{Chen2025ecg5000}, and ETTh1~\cite{haoyietal2021etth1} datasets. 

We assess the capabilities of \systemName along three dimensions aligned with its design goals: (i) usability and interface simplicity; (ii) architectural adaptability across tasks and components; and (iii) performance benchmarking. Together, these evaluations aim to characterize the extent to which \systemName facilitates reproducible, modular experimentation in the TSFM landscape.

\vspace{-0.5em}
\subsection{Usability and Interface Simplicity}

 \systemName prioritizes usability through a set of intuitive APIs. Users can compose, fine-tune and execute complex pipelines via a small number of method calls, abstracting away internal implementation details. To validate \systemName, we first replicate the PPG-based physiological monitoring benchmark from the PaPaGei repository~\cite{pillai2024papagei}, which includes tasks such as predicting systolic/diastolic BP and heart rate. As shown in Table~\ref{table:accuracy_comparison}, our standardized pipeline matches the baseline results, achieving an error rate within 1\% of the original implementation across all prediction tasks and FM backbones. In addition, \systemName simplifies the exploration of novel architectural combinations. For instance, constructing a complex pipeline that combines the Moment backbone with a custom encoder, an MLP decoder, and a LoRA adapter for fine-tuning requires only seven lines of code (see Appendix \ref{appA} for details). This demonstrates the toolkit's core design principle: new components can be implemented by sub-classing simple abstract classes and are immediately interoperable within the \texttt{Pipeline} interface, enabling rapid and systematic experimentation.

\vspace{-0.5em}
\subsection{Architectural Adaptability}
\begin{table}[t]
\centering

\begin{minipage}{0.45\textwidth}
\centering
\scriptsize
\renewcommand{\arraystretch}{1.5}

\begin{tabular}{lccc}
\toprule
\textbf{Decoder}    & \textbf{Chronos} & \textbf{Moment} & \textbf{PaPaGei-S} \\ \toprule
SVM                 & 93.15            & 94.02           & 89.86              \\
KNN                 & 91.64            & 92.71           & 86.91              \\
LR & 58.37            & 92.02           & 89.51              \\
RF      & 91.26            & 91.80           & 87.53              \\
MLP                 & 90.26            & 93.00           & 75.55              \\ \bottomrule
\end{tabular}
\label{tab:variety_decoder}
\end{minipage}%
\hfill
\begin{minipage}{0.54\textwidth}
\centering
\scriptsize
\begin{tabular}{lcccc}
\toprule
                     & \multicolumn{2}{c}{\textbf{\begin{tabular}[c]{@{}c@{}}HC Task\end{tabular}}} & \multicolumn{2}{c}{\textbf{\begin{tabular}[c]{@{}c@{}}EF Task\end{tabular}}} \\ \cmidrule(lr){2-3} \cmidrule(lr){4-5}
\textbf{Metric}      & Base & FMTK & Base & FMTK \\ \midrule
\textbf{Time (s)}  &      &      &      &      \\
Finetune             & 10.83& 11.15& 50.38& 50.20 \\
Predict              & 0.03 & 0.03 & 0.04 & 0.04 \\ \midrule
\textbf{Peak Memory (MB)} &      &      &      &      \\
Finetune             & 563.11&569.17&804.86&779.08 \\
Predict              & 562.62&562.62&797.07&740.80 \\ \midrule
\textbf{Energy (J)} &      &      &      &      \\
Finetune             & 892.52& 923.19&16718.41&16689.45 \\
Predict              & 7.84&8.14&16.56&16.86 \\ 
\bottomrule
\end{tabular}
\label{tab:performance_comparison}
\end{minipage}
\vspace{1em}\caption{\textbf{(left)} Accuracy of Chronos-large, Moment-large, and PaPaGei-S backbone with multiple decoders for Heartbeat Classification task. Table 3: \textbf{(right)} Performance comparison of \systemName using the Moment-base backbone with SVM and MLP decoders for Heartbeat Classification (HC) and Energy Forecasting (EF), respectively. Fine-tuning time is measured over the entire train dataset, while prediction time is measured per batch.}
\vspace{-2em}
\end{table}

One of the central design goals of \systemName is to enable modular experimentation across different architectural choices. To demonstrate this, we first show how a single backbone can be paired with various decoders. As shown in Table 2, the Chronos, Moment and PaPaGei backbone can be seamlessly combined with diverse decoders such as an SVM, MLP, KNN, Logistic Regression (LR) or Random Forest (RF) for Heartbeat Classification task. Furthermore, \systemName simplifies the process of repurposing the entire pipeline for different tasks and backbones with minimal code changes. For instance, Table~\ref{table:accuracy_comparison} presents results for both Heartbeat Classification and Energy Forecasting tasks using the Chronos, Moment and PaPaGei backbones, all implemented within our unified toolkit using non-traditional components. \systemName successfully accommodates a rich diversity of components and tasks, fulfilling its core design goal of composability. Notably, the table demonstrates novel compositions, even beyond default use-cases, as in the case of Chronos and PaPaGei. 

\vspace{-0.5em}
\subsection{Performance Benchmarking}
To validate that the modularity and ease of use in \systemName come at an acceptable computational cost, we conduct an extensive performance analysis comparing our toolkit with the baseline implementations. We used Moment-Base model with SVMDecoder for Heartbeat Classification, and with MLPDecoder for Energy Forecasting task. As shown in Table 3, \systemName incurs a minimal $\sim$3\% overhead for fine-tuning and prediction time across both tasks. It achieves a $\sim$7\% reduction in peak inference GPU memory for the Energy Forecasting task, with negligible memory overhead during training. Furthermore, the toolkit exhibits an energy consumption comparable to the baseline implementation.

\section{Conclusion}
This work introduces \systemName, a lightweight yet extensible toolkit for modularizing, composing, and benchmarking time series foundation model pipelines. By decoupling architectural components and unifying evaluation semantics, \systemName lowers the barrier to rigorous, reproducible experimentation across a rapidly expanding design space. While instantiated for time series tasks, the framework is broadly applicable to any foundation model workflow exhibiting an encoder–backbone–decoder structure, offering a template for systematic evaluation across modalities. For future work, we are expanding this toolkit to support other types of foundation models, adapters and add support for runtime optimizations. 

\newpage
\begin{ack}

We thank the anonymous reviewers for their valuable insight and feedback. This research is supported by National Science Foundation (NSF) grants 2211301, 2211302, 2325956, 2213636, 2211888, NIH grant 1P41EB028242 and US Army grant W911NF-17-2-0196.
\end{ack}
\bibliographystyle{plain}
\bibliography{main}

\appendix
\newpage
\part*{Appendices}

\section{Implementation Details}\label{appA}
\subsection{Code Example for a Modular Pipeline}
The following code Listing~\ref{lst:moment_example} demonstrates how \systemName can be used to construct a complex pipeline. It loads the Moment backbone, adds a custom encoder, attaches an MLP decoder, and applies a LoRA adapter for the heart rate prediction task using the PPG-BP dataset.
\begin{listing}[ht]
\centering
\begin{minipage}{\linewidth}
\begin{minted}[fontsize=\small, linenos]{python}
lora_config = LoraConfig(r=64, lora_alpha=32, target_modules=["q", "v"], lora_dropout=0.05)
P=Pipeline(MomentModel(model))
P.add_decoder(MLPDecoder(cfg={'input_dim':1024,'output_dim':1,'hidden_dim':128},load=True)
P.add_encoder(LinearChannelCombiner(cfg={num_channels=3,new_num_channels=1}, load=True)
P.add_adapter(lora_config)
P.train(dataloader_train,parts_to_train=['encoder','decoder','adapter'],cfg=task_cfg['train_config'])
y_test,y_pred=P.predict(dataloader_test,cfg=task_cfg['inference_config'])
\end{minted}
\end{minipage}\hfill
\vspace{1em}
\centering\caption{Example to setup pipeline abstraction attaching Moment Backbone with MLP decoder, Linear channel encoder and LORA adapter for heart rate prediction task using PPG Dataset.}
\label{lst:moment_example}
\end{listing}

\subsection{Standardized Interface for Integrating Customized Components}
\begin{listing}[ht]
\centering
\begin{minipage}{0.46\linewidth}
\begin{minted}[fontsize=\small]{python}
class BaseModel:
    def __init__(self, cfg):
        """
        Loading model
        """
        
    def preprocess(self, batch):
        """
        Match the shape and preprocess 
        before sending it to model.
        Args:
            batch
        Returns: 
            batch
        """
        
    def postprocess(self,embedding):
        """
        Postprocess the embedding to 
        standard shape for next component
        Args:
            embedding
        Returns: 
            embedding 
        """    
        
    def forward(self, batch):
        """
        Method for forward pass for 
        one batch
        Args:
            batch
        Returns: 
            embedding 
        """
        ...
    
    def trainable_parameters(self):
        """
        Get trainable paramters 
        out of model
        Returns: 
            Iterable[torch.nn.Parameter]
        """   
        ...
\end{minted}
\end{minipage}\hfill
\begin{minipage}{0.49\linewidth}
\begin{minted}[fontsize=\small]{python}
# Encoder
class LinearChannelEncoder(BaseModel):
    def __init__(self,cfg):
        """
        Dimentionality reduction using 
        linear layer from number of 
        input channel to 
        number of output channel.
        """
        
    def preprocess(self, batch):
        """
        Preprocessing it to acceptable
        input dimension [B,C,L]
        """

    def postprocess(self,embedding):
        """
        Postprocess the embedding to 
        standard shape for 
        foundation model [B,C,L]
        """    
\end{minted}

\vspace{0.6em}

\begin{minted}[fontsize=\small]{python}
# Backbone wrapper
class ChronosBackbone(BaseModel):
    def __init__(self, cfg):
        """
        Loading Chronos
        """
        ...
    def preprocess(self, batch):
        """
        Preprocessing it to acceptable
        input dimension [B*C,L]
        """
        ...
    def forward(self, batch):
        """
        One forward pass through
        preprocess, backbone and 
        postprocess
        """
        ...
    def postprocess(self, embedding):
        """
        Postprocess the embedding to 
        standard shape for 
        decoder [B,C,T,L] | [B,C,L]
        """ 
        ...
\end{minted}

\vspace{0.6em}

\begin{minted}[fontsize=\small]{python}
# Decoder
class MLPDecoder(BaseModel):
    def __init__(self, cfg):
        ...
    def preprocess(self, batch):
        """
        Preprocessing it to acceptable
        input dimension [B,L] 
        based on cfg
        """
        ...
\end{minted}
\end{minipage}

\caption{\textbf{(left)}  Common interface description for encoder, backbone and decoders. \textbf{(right)} Example illustrating the pre-processing and post-processing steps involved in integrating a Linear channel encoder, Chronos backbone, and MLP decoder into a unified pipeline.
}
\label{lst:pipeline}
\end{listing}
To integrate new components and enable modular composition in the time-series foundation-model (TSFM) pipeline, we adopt a single, minimal interface shared by encoders, backbones, and decoders (Listing \ref{lst:pipeline} (left)). Each component minimally implements the BaseModel interface: preprocess adapts inbound tensors to the component's expected format (shape/device/dtype/reduction) and postprocess standardizes the outbound representation for the next stage. This separation isolates model-specific code from pipeline composition, allowing components to be swapped without code changes in the pipeline abstraction. For example as shown in Listing \ref{lst:pipeline} (right), the Chronos backbone receives inputs shaped [B,C,L]. Internally, the model operates on a flattened view [B*C,L], which is produced in the preprocess. Chronos returns embeddings [B,E,L] (where E is the token/feature dimension); postprocess reshapes these to [B,C,E,L] so that downstream decoders (e.g., the MLP decoder in Listing \ref{lst:pipeline} (right)) can consume a consistent, a channel aware representation. This standardized boundary ensures that alternative encoders/backbones/adapters/decoders can be composed interchangeably within the same pipeline.

\end{document}